\documentclass[10pt,twocolumn,letterpaper]{article}

\usepackage[pagenumbers]{cvpr} % To force page numbers, e.g. for an arXiv version

\usepackage{graphicx}
\usepackage{amsmath}
\usepackage{amsthm}
\usepackage{amssymb}
\usepackage{booktabs}
\usepackage{multicol}
\usepackage{multirow}
\usepackage{subcaption}
\usepackage{multirow}
\usepackage{verbatim}
\usepackage{bbding}
\usepackage{color}
\usepackage{algorithm}
\usepackage{algorithmic}
\usepackage{dsfont}

\usepackage[pagebackref,breaklinks,colorlinks]{hyperref}

\usepackage[capitalize]{cleveref}
\crefname{section}{Sec.}{Secs.}
\Crefname{section}{Section}{Sections}
\Crefname{table}{Table}{Tables}
\crefname{table}{Tab.}{Tabs.}

\def\cvprPaperID{2846} % *** Enter the CVPR Paper ID here
\def\confName{CVPR}
\def\confYear{2022}

\begin{document}

%%%%%%%%% TITLE - PLEASE UPDATE
\title{AKB-48: A Real-World Articulated Object Knowledge Base}
%\title{AKB-48: An Articulated Object Knowledge Base for Real-World Articulation Analysis}

%AKB-48: A Real-World Model Repository and Benchmark for Articulated Object Analysis}

\author{Liu Liu, Wenqiang Xu, Haoyuan Fu, Sucheng Qian, Yang Han, Cewu Lu\\
Shanghai JiaoTong University \\
%\quad bbb University\\
{\tt\small \{liuliu1993, vinjohn, simon-fuhaoyuan, qiansucheng, lucewu\}@sjtu.edu.cn}
% For a paper whose authors are all at the same institution,
% omit the following lines up until the closing ``}''.
% Additional authors and addresses can be added with ``\and'',
% just like the second author.
% To save space, use either the email address or home page, not both
}

% \author{Liu Liu\\
% Institution1\\
% Institution1 address\\
% {\tt\small firstauthor@i1.org}
% % For a paper whose authors are all at the same institution,
% % omit the following lines up until the closing ``}''.
% % Additional authors and addresses can be added with ``\and'',
% % just like the second author.
% % To save space, use either the email address or home page, not both
% \and
% Second Author\\
% Institution2\\
% First line of institution2 address\\
% {\tt\small secondauthor@i2.org}
% }

\twocolumn[{%
\renewcommand\twocolumn[1][]{#1}%
\maketitle
\begin{center}
    \centering
    \captionsetup{type=figure}
    %\fbox{\includegraphics[width=\linewidth]{figures/abstract12.png}}
    \includegraphics[width=\linewidth]{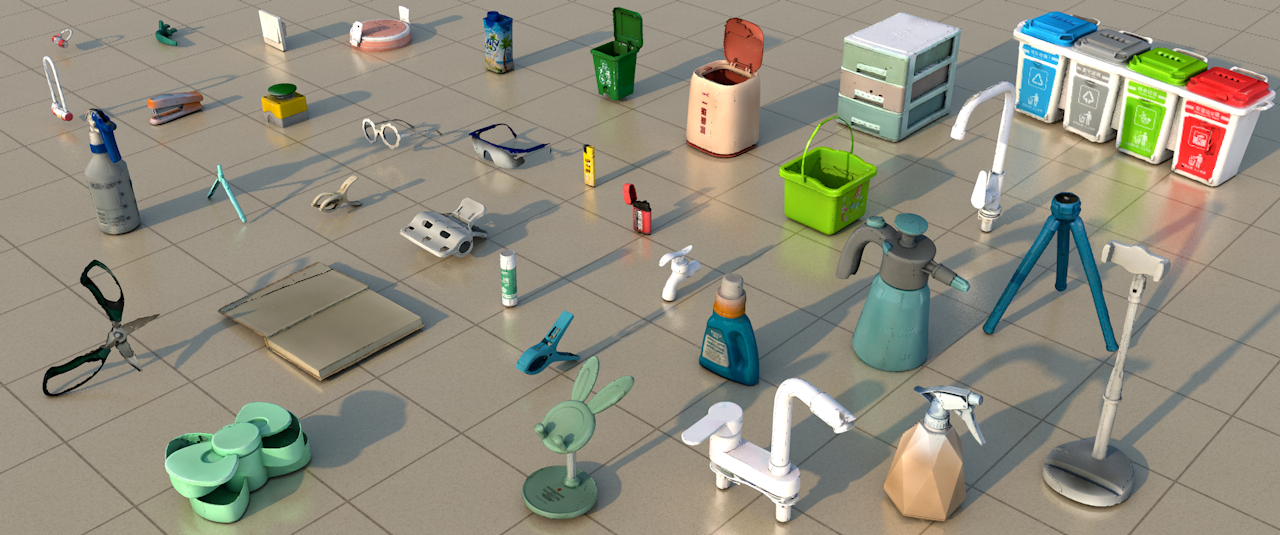}
    \captionof{figure}{AKB-48 consists of 2,037 articulated object models of 48 categories scanned from the real world. The objects are annotated with ArtiKG, and can support a full task spectrum from computer vision to robotics manipulation.}
    \label{fig:abstract}
\end{center}%
}]

%%%%%%%%% ABSTRACT
\begin{abstract}
    Human life is populated with articulated objects. A comprehensive understanding of articulated objects, namely appearance, structure, physics property, and semantics, will benefit many research communities. As current articulated object understanding solutions are usually based on synthetic object dataset with CAD models without physics properties, which prevent satisfied generalization from simulation to real-world applications in visual and robotics tasks. To bridge the gap, we present AKB-48: a large-scale \textbf{A}rticulated object \textbf{K}nowledge \textbf{B}ase which consists of 2,037 real-world 3D articulated object models of \text{48} categories. Each object is described by a knowledge graph \text{ArtiKG}. To build the AKB-48, we present a fast articulation knowledge modeling (FArM) pipeline, which can fulfill the ArtiKG for an articulated object within 10-15 minutes, and largely reduce the cost for object modeling in the real world. Using our dataset, we propose AKBNet, a novel integral pipeline for Category-level Visual Articulation Manipulation (C-VAM) task, in which we benchmark three sub-tasks, namely pose estimation, object reconstruction and manipulation. Dataset, codes, and models will be publicly available at \url{https://liuliu66.github.io/articulationobjects/}.
    %we establish three benchmark tasks for evaluating real-world articulation analysis: category-level articulated object pose estimation, articulated object reconstruction and articulated object manipulation. We also propose baseline methods for these tasks and demonstrate their performance. Dataset, codes, and models will be public available.
\end{abstract}
%%%%%%%%% BODY TEXT
\section{Introduction}
\label{sec:intro}

Articulated objects, composed of more than one rigid part connected by joints allowing rotational or translational movements in 3D space, are pervasive in our daily life. Knowledge about the articulated objects can be beneficial to many research communities, such as computer vision, robotics and embodied AI. Thus, many articulated object datasets have been proposed to facilitate the research, such as PartNet-Mobility\cite{xiang2020sapien}, ReArt-48\cite{liu2021towards}, RBO\cite{martin2019rbo}. However, these datasets generally focus more on the structural information (\eg part segmentation, kinematic structure), but pay less attention to the appearance (\eg texture, fine geometry), the physics properties (\eg per-part mass, inertial, material and friction) and semantics (\eg category, affordance). While some important tasks heavily rely on these information such as object detection (\textit{texture}) \cite{borrego2018applying}, 3D reconstruction (\textit{fine geometry}) \cite{maier2017intrinsic3d}, object manipulation (\textit{physics property}) \cite{chang2020sim2real2sim}, and so on, the lacking of such object knowledge in these datasets can prevent satisfied generalization for the learning models. % 

\begin{figure*}[t]
    \centering
    \includegraphics[width=\linewidth]{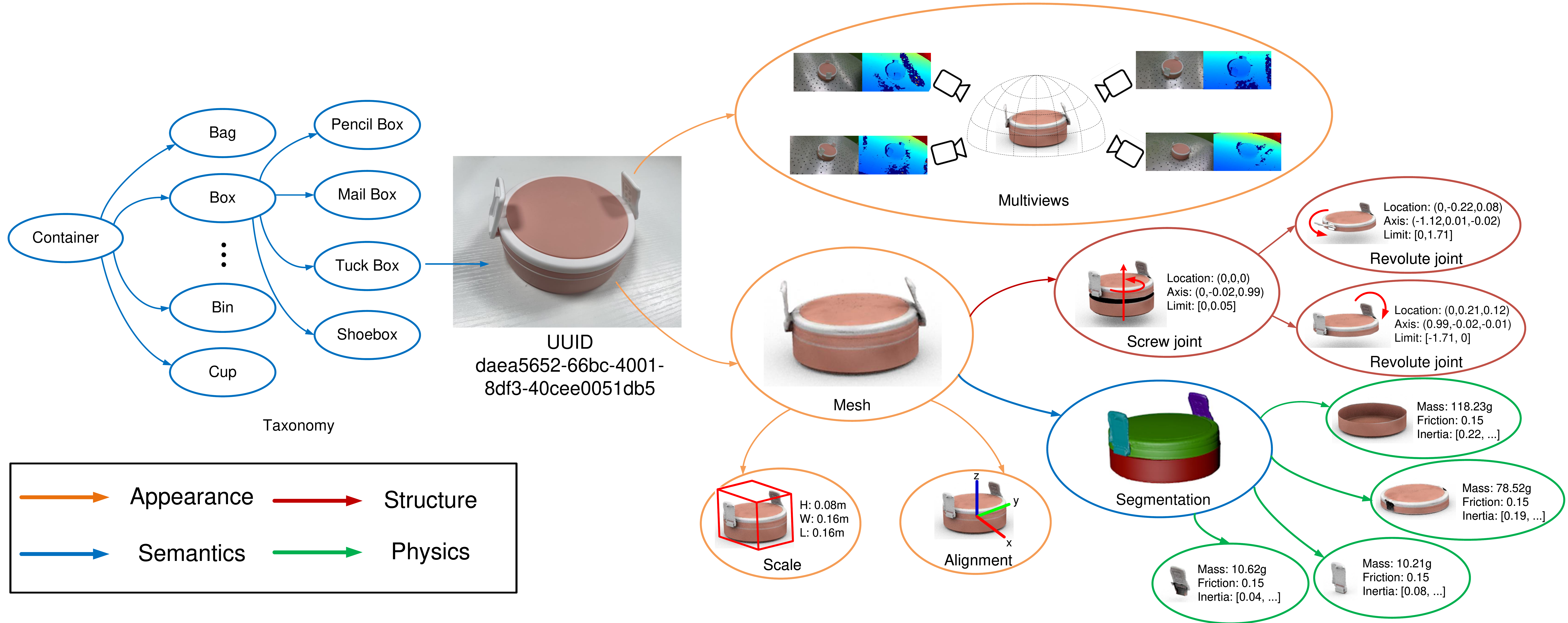}
    \caption{The Articulation Knowledge Graph (ArtiKG) defined in AKB-48 dataset. In ArtiKG, we annotate four types of knowledge: Appearance, Structure, Semantics and Physics property. The values are rounded up to percentile in this figure for presentation. }
    \label{fig:object_kg}
\end{figure*}

\begin{figure}[t]
    \centering
    \includegraphics[width=\linewidth]{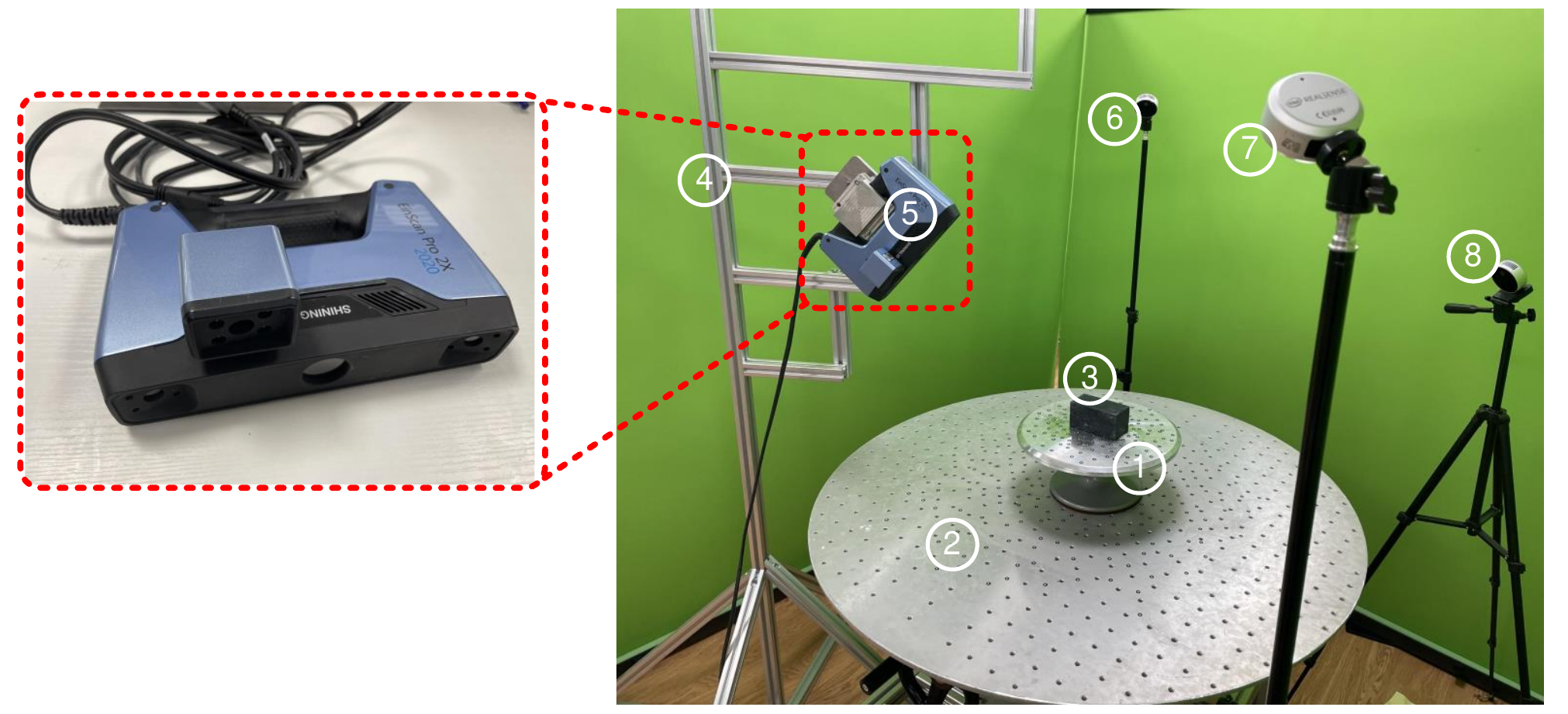}
    \caption{The task-specific model acquisition equipment. (a) 1 is a Rotating turntable for objects with multiple scales. 2 is a tracking marker. 3 is a light-absorbing item. 4 is a lift bracket. 5 is the Shining 3D scanner. 6-8 are the realsense L515 cameras for capturing multiviews of objects.}
    \label{fig:model_acq_equipment}
\end{figure}

To boost the research on articulated objects, in this paper, we present \textbf{AKB-48}: a large-scale real-world \textbf{A}rticulated \textbf{K}nowledge \textbf{B}ase which includes 48 categories, 2,037 instances. For each instance, the object model is scanned from the real counterpart and refined manually (Sec. \ref{sec:model_acq}), and the object knowledge is organized to a graph, named \textbf{Art}iculation \textbf{K}nowledge \textbf{G}raph (\textbf{ArtiKG}), which contains the detailed annotations of different kinds of object attributes and properties (Sec. \ref{sec:artikg}). To make the scanning and annotation process feasible for large datasets, we present a \textbf{F}ast \textbf{Ar}ticulation Knowledge \textbf{M}odeling (\textbf{FArM}) pipeline (Sec. \ref{sec:farm}). In detail, we develop an object recording system with 3D sensors and turntables, a GUI that integrates structural and semantic annotations, and standard real-world experiments for physics property annotation (Fig. \ref{fig:model_acq_equipment}). In this way, we can save a large amount of money and time budget for modeling real-world articulated objects ($\sim$\$3 to buy, 10-15min to annotate per object). A thorough comparison between the CAD modeling and reverse scanning can be referred to Sec. \ref{sec:model_acq}. To summarize, our pipeline can save 33 folds on the money budget and 5 folds on the time budget.

To utilize the AKB-48 for research, we propose \textbf{AKBNet}, a novel integral pipeline for \textbf{C}ategory-level \textbf{V}isual \textbf{A}rticulation \textbf{M}anipulation (\textbf{C-VAM}) task. To address C-VAM task, the vision system AKBNet should be able to estimate the object pose, reconstruct the object geometry and reason the policy for manipulation at category level. Thus, it consists of three perception sub-modules:
%To utilize the ABK-48 for research, we also propose a deep learning framework, named \textbf{A}rticulation \textbf{P}erception and \textbf{M}anipulation \textbf{Net}work (\textbf{AKBNet}), to perform real-world articulated object visual analysis and physical manipulation. Specifically, the AKBNet consists of three key components: category-level articulated object pose estimation, articulated object reconstruction and articulated object manipulation.

%To utilize the AKB-48 for research, we also provide benchmarks for different articulated object analysis tasks. Specifically, the tasks consist of category-level articulated object pose estimation, articulated object reconstruction and articulated object manipulation.

\begin{itemize}
    \item \textbf{Pose Module} for \textit{Category-level Articulated Object Pose Estimation}. This module aims to estimate the per-part 6D pose of an unseen articulated object in one category. However, prior researches generally study on \textit{kinematic category}, that is objects of a category are defined to have the same kinematic structure. Our pose module extends the concept of ``category'' to \textit{semantic category}, in which the category is defined by the semantics and different kinematic structures are allowed. (Sec. \ref{sec:pose_estimation})
    %We extend the problem setting \textit{semantic category}, in which different kinematic structures are allowed. In this work, we generate a dataset by rendering a large number of RGB-D images with various scenes and validate our general version of A-NCSH\cite{li2020category} performance on real-world. 
    \item \textbf{Shape Module} for \textit{Articulated Object Reconstruction}. After the pose is obtained, along with the shape code encoding from input images, we can reconstruct the shape for each part \cite{mu2021sdf}. Full geometry is critical for manipulation to determine where to interact with. (Sec. \ref{sec:object_reconstruction})
    \item \textbf{Manipulation Module} for \textit{Articulated Object Manipulation}. Once we obtain the articulation information (\eg part segments, per-part pose, joint properties, full mesh, etc.) through perception, we can reason the interaction policy over the observations. We benchmark manipulation tasks with opening and pulling that are corresponding to revolute and prismatic joint respectively. (Sec. \ref{sec:articulation_manipulation}) % by training a reinforcement learning agent
\end{itemize}

To evaluate the AKBNet, we report the results individually and systematically. For individual evaluation of each module, we assume the input to the module is the ground truth of the last module, while for systematical evaluation, the input is the output of the last module. Apparently, we cannot benchmark all the tasks which can be supported by the proposed AKB-48. We hope it could serve as a good platform for future articulation research in computer vision and robotics community. %For category-level articulated object pose estimation task, we report Average Precision (AP) under rotation, translation and 3D IoU metrics. For object reconstruction task, we report Chamfer-L1 distance. For manipulation task, we report successful rate under 100 rollouts.  
%Apparently, we cannot evaluate all the tasks which can be supported by the proposed AKB-48 in one framework even though our AKBNet benchmarks the popular visual articulation analysis tasks as well as one of the important robotic-object interaction tasks. We hope AKB-48 could serve as a good platform for future articulation research in computer vision and robotics community.

Our contributions can be summarized in three folds:

\begin{itemize}
  \item We introduce AKB-48, containing 2,037 articulated models across 48 categories, in which we adopt a multi-modal knowledge graph ArtiKG to organize the rich annotations. It can help mitigate the gap between the current vision and embodied AI researches. To the best of our knowledge, it is the first large-scale articulation dataset with rich annotations collected from the real world.
  
  \item We propose a fast articulation knowledge object modeling pipeline, FArM, which makes it much easier to collect articulated objects from the real world. Our pipeline greatly eases the cost on time and money when building real-world 3D model datasets.
  
  \item We propose a novel pipeline AKBNet for the integral category-level visual articulation manipulation (C-VAM) task. Experiments show our approach is effective both individually and systematically in the real world.
  %\item We propose an articulation analysis framework AKBNet to provide three major benchmarks for articulation object analysis with our ABK-48. Experiments illustrate the usefulness and superiority of AKB-48 for learning models performance on the real world.
  %\item We benchmark three state-of-the-art algorithms for articulation object analysis with our AKB-48. Experiments illustrate the usefulness and superiority of AKB-48 for learning models performance on the real world. 
\end{itemize}

\section{Related Work}

\paragraph{3D Model Repositories and Datasets.} An unavoidable challenge for analyzing 3D objects, especially for articulated objects, is the lack of large-scale training data with sufficient 3D models and full annotations. To the best of our knowledge, current 3D model repositories prefer to collect CAD models by searching from the Internet such as Trimble 3D Warehouse and Onshape \cite{koch2019abc}. ShapeNet \cite{chang2015shapenet} collects approximately 3 million shapes from online model repositories and categorizes them based on WordNet \cite{miller1995wordnet} taxonomy. But although ShapeNet contains many articulated categories, the models of ShapeNet can only be considered as rigid shapes since they do not define parts within them. To deal with this problem, Mo et al. \cite{mo2019partnet} first present a large-scale dataset PartNet that annotates hierarchical part semantic segmentation based on a subset of ShapeNet \cite{chang2015shapenet}. One critical problem in PartNet is that it pays much attention to labeling each semantic part but ignores the kinematics structures. To solve this issue, PartNet-Mobility \cite{xiang2020sapien} and Shape2Motion \cite{wang2019shape2motion} further annotate joint properties on the shapes, which target at articulation research. 

% But ShapeNet only focuses on rigid shapes without articulation or fine-grained annotations. Based on ShapNet, Mo et al. \cite{mo2019partnet} selects around 20K models and annotates hierarchical part semantic segmentation, which first presents part-level 3D recognition tasks. One of the problems of PartNet is that they pay much attention on segmenting each rigid part but ignore the motion properties among these parts. To solve this issue, PartNet-Mobility \cite{xiang2020sapien} and Shape2Motion \cite{wang2019shape2motion} further refine the annotations and present articulation model repositories in computer vision community.

These datasets mostly follow the model construction paradigm from ShapeNet: collecting CAD models from the Internet and providing specific annotations for different tasks. This allows the early works (ShapeNet \cite{chang2015shapenet}, ABC dataset \cite{koch2019abc}, etc.) to quickly build large-scale object model bases. However, when the task is required to investigate new categories or kinematic structures, artists need to manually build proper CAD models from scratch, which is very time-consuming and laborious. On the other hand, current real-world researches focus on instance-level tasks so they tend to build small-scale model datasets such as YCB \cite{calli2015ycb} and RBO \cite{martin2019rbo}. Therefore, the data volume makes it hard to be adopted in our category-level articulation tasks, which requires generalization capacity among different instances. In this paper, we present AKB-48 as the first large-scale real-world base for articulation analysis.

%Most of above datasets follow the model construction paradigm from ShapeNet: collecting shapes from Internet and providing specific annotations for different visual tasks. This limits the scalability when investigating new categories or kinematic structures of articulated objects. On the other hand, high labor cost also prevents real-world model repository and dataset building \cite{calli2015ycb}. Although Martın-Martın et al. \cite{martin2019rbo} publish a small-scale real-world articulation object dataset, the data volume makes it hard to be adopted in real-world applications. In this paper, we present AKB-48 as the first large-scale real-world dataset with rich annotations.

%with an efficient articulated object modeling pipeline.
%{\color{red} really think again why ordinary robot object dataset are small. The reason why previous robot community did not build a large dataset is due to the immaturity of computer vision techniques, so there is no point to build a category-level dataset.}

\paragraph{Articulation-related Tasks.} Articulated objects have been investigated for decades in both vision and robotics communities but hold different emphases. In vision tasks, current works tend to solve category-level object recognition, segmentation or pose estimation that focus on generalization among objects. Yi et al. \cite{yi2019deep} take a pair of unsegmented shape representations as input to predict part segmentation and deformation. For tackling with unseen objects, Li et al. \cite{li2020category} follow the pose estimation setting and propose a normalized coordinate space to estimate 6D pose and joint state for articulated objects. In terms of joint-centered perception tasks, several works attempt to mine joint configurations of articulated objects \cite{jain2020screwnet, liu2020nothing, zeng2020visual}. To investigate manipulation points for articulated objects from visual input, Mo et al. attempt to define six types of action primitives and predict interactions \cite{mo2021where2act}. In terms of robotics community, researchers usually solve interaction or manipulation tasks to achieve articulation inference such as robot interactive perception \cite{katz2008manipulating}, feedback by visual observation \cite{hausman2015active} and task integration \cite{martin2016integrated}. Besides, some works attempt to bridge the gap between vision and manipulation but still suffer from the small-scale issue. Therefore, we propose AKBNet to deal with category-level articulation tasks.

% There are two major types of tasks that target at solving articulation object problem: vision task and manipulation \& interaction task. Yi et al. \cite{yi2019deep} take a pair of unsegmented shape representations as input to predict part segmentation and deformation. For tackling with unseen objects, Li et al. \cite{li2020category} follow the pose estimation setting and propose a normalized coordinate space to estimate 6D pose and joint state for articulated objects. In terms of joint centered perception task, several works attempt to mine joint configurations of articulated objects \cite{jain2020screwnet, liu2020nothing, zeng2020visual}. To investigate how to manipulate or interact with articulated objects, Mo et al. define six types of action primitives and predict interactions from visual input. Besides, many works focus on robot-object-manipulation ways to achieve articulation inference such as robot interactive perception \cite{katz2008manipulating}, feedback by visual observation \cite{hausman2015active} and task integration \cite{martin2016integrated}. In this paper, we benchmark four articulation related tasks based on our AKB-48 and validate the performance on the real world.

\section{Articulation Knowledge Base, AKB-48}\label{sec:dataset}
When constructing the knowledge base, three instant questions should be answered: (1) What kinds of knowledge should we annotate on the object? (2) What objects should we annotate, those from the real or the simulated world? (3) How to annotate the object knowledge efficiently?
To answer these questions, we describe the ArtiKG in Sec. \ref{sec:artikg}, make a thorough discussion on the object selection in Sec. \ref{sec:model_acq}, and finally propose the FArM pipeline in Sec. \ref{sec:farm} and provide analysis (diversity, difficulty) about the dataset in Sec. \ref{sec:dataset_analysis}.

\subsection{Articulated Object Knowledge Graph, ArtiKG}\label{sec:artikg}
Different tasks require different kinds of object knowledge, to unify the annotation representation, we organize it into a multi-modal knowledge graph, named ArtiKG. The ArtiKG consists of four major parts, namely appearance, structure, physics, and semantics. The details are described in the following and visualized in Fig. \ref{fig:object_kg}.

\vspace{0.5em}\noindent\underline{Appearance.} For each instance, we store its shape with mesh data structure along with the textures. When scanning the object from the real world, we also collect the multi-view RGB-D snapshots of the object.

\vspace{0.5em}\noindent\underline{Structure.} The key difference between the articulated object and the rigid object is the kinematic structure. The articulated object has concepts like joint and part, which are not meaningful for the rigid object. For each joint, we annotate the joint type, parameters, and movement limits. For each part, we segment each kinematic part.

\vspace{0.5em}\noindent\underline{Semantics.} After the basic geometric and structural information is annotated, we begin to assign the semantic information to the object in a coarse-to-fine process. We give a uuid to each instance. Then we assign the category and the corresponding taxonomy to the object according to WordNet \cite{miller1995wordnet}. We also label the semantic part. Though we already annotate the kinematic part, it is not quite the same as the semantic part. Take a mug with a handle, for example, the handle is not attached to the mug body through a joint, thus it is not a kinematic part, but it is a semantic part as it indicates where the human normally grabs the mug. 

\vspace{0.5em}\noindent\underline{Physics property.} Real objects exist in the physical world and typically have physical properties, which are important for accurate simulation and real-world manipulation \& interaction on articulated objects. Thus, we store physical attribute annotations for our models, involving per-part mass, Per-part inertial, material and surface friction.

\vspace{0.5em}\noindent\textbf{Discussion.} In this section, we only describe the object knowledge thet should take human's effort to annotate, for those which can be calculated through algorithms or trivially inferred like surface normal, collision mesh/simplified mesh, intrinsic dimensions, are not discussed. Besides, as the annotation information is modular organized, it is convenient for new attributes to be added to the ArtiKG. Besides, though the ArtiKG is designed for articulated objects, it can also be trivially extended to rigid, and flexible objects.

\subsection{Object Selection: Real-world Scanning v.s CAD Modeling}\label{sec:model_acq}
 
The choice between real-world scanning and CAD modeling are considered from two perspectives, namely annotation accuracy, cost on time and money.

\vspace{0.5em}\noindent\textbf{Annotation Accuracy.} According to the content of the ArtiKG, we can see objects from the real world have multiple advantages over the CAD models, such as appearance and physics property. But admittedly, the CAD model can model inner structures such as the GUNdam or the transformer, while scanning techniques focus more on the surface. Since such objects with inner structures that cannot be easily disassembled posit challenges for both artists and the scanners, we would like to update these objects when techniques are more ready. Fortunately, most daily objects can be disassembled, so the scanning techniques can properly handle them. 

\vspace{0.5em}\noindent\textbf{Cost on Time and Money.} As discussed earlier, ShapeNet-like model collection paradigm is limited to large time and money cost of artists' manual CAD model building when investigating new categories or kinematic structures. On the other hand, many daily articulated objects are cheap in reality and can be scanned by a layman. We compare the average money and time budget in Table \ref{tab:budget}. For CAD modeling, it is estimated from outsourcing services in Taobao website\footnote{\url{https://www.taobao.com}}. From our survey, most artists spend more than 2 hours (over 120 minutes) to model an articulated object and the labor cost is averagely over 100 dollars for one.

%As mentioned earlier, rigid object datasets inherits a large amount of asset properties, the CAD models from the Internet. However, when it comes to the articulated object, the existing models with movable joints are hard to find, mostly due to the fact that it is much more time-consuming for an artist to build a proper CAD model for daily articulated objects. While on the other hand, many daily articulated objects are cheap in reality and can be scanned by a layman. We compare the average money and time budget in Table \ref{tab:budget}. For CAD modeling, it is estimated from outsourcing services in Taobao website\footnote{\url{https://www.taobao.com}}. From our survey, most artists spend more than 2 hours (over 120 minutes) to model an articulated object and the labor cost is averagely over 100 dollars for one.

\begin{table}[ht!]
    \centering
    \begin{tabular}{c|c|c}
    \hline
         &  CAD modeling & Real-world Scanning \\
         \hline
        Time (min) & $>$120 & 20 \\
        Money (\$) & $>$100 & 3 \\
    \hline
    \end{tabular}
    \caption{Budget comparison between our real-world scanning and CAD modeling for articulated objects.}
    \label{tab:budget}
\end{table}

To note, we are aware that many important articulated objects in the real world are rather expensive like laptop, microwave oven, doors etc. In such cases, we either collect only the ones we can collect from the homes without re-buying, or buy one to measure the basic information and propagate to the existing simulated objects like in PartNet-Mobility \cite{xiang2020sapien}. For these objects, the ArtiKG is labeled as ArtiKG-sim.

%\tabincell{c}{Avg. \\ \# vertices}
\begin{table*}[tbh]
%\scriptsize
\centering
\resizebox{0.8\linewidth}{!}{
\begin{tabular}{l|ccc|cc|cc|ccc}
\hline
\multirow{2}{*}{Dataset} & \multicolumn{3}{c|}{Appearance} & 
\multicolumn{2}{c|}{Structure} & \multicolumn{2}{c|}{Semantics} & \multicolumn{3}{c}{Physics} \\
\cmidrule(r){2-11}
& Num & AV & AT & Part & Joint & ST & PS & PM & PI & PF \\
\hline
\textit{Synthetic Model Dataset}\\
ShapeNet\cite{chang2015shapenet} & $>$50K & $<$2K & $<$5K & - & - & \checkmark & - & - & - & - \\ 
PartNet\cite{mo2019partnet} & $>$20K & $<$2K & $<$5K & \checkmark & - & \checkmark & - & - & - & - \\
Shape2Motion\cite{wang2019shape2motion} & 2K & $<$0.5K & $<$1K & \checkmark & \checkmark & - & - & - & - & - \\
PartNet-Mobility\cite{xiang2020sapien} & 2K & $<$0.5K & $<$1K & \checkmark & \checkmark & \checkmark & \checkmark & - & - & - \\
\midrule
\textit{Real-World Model Dataset}\\
YCB\cite{calli2015ycb} & 21 & $\sim$40K & $\sim$90K & - & - & - & - & - & - & - \\
LineMod\cite{2012Multimodal} & 15 & $\sim$19K & $\sim$39K & - & - & - & - & - & - & - \\
RBO\cite{martin2019rbo} & 14 & $\sim$5K & $\sim$10K & \checkmark & \checkmark & - & - & - & - & - \\
AKB-48(Ours) & 2,037 & $\sim$63K & $\sim$126K & \checkmark & \checkmark & \checkmark & \checkmark & \checkmark & \checkmark & \checkmark \\
\hline    
\end{tabular}}
\caption{Comparison with other popular model datasets. Our AKB-48 dataset provides four types of information for rich annotations in our ArtiKG: Appearance, Structure, Semantics and Physics. \textbf{AV: }Average number of vertices. \textbf{AT: }Average number of triangles. \textbf{ST: }Semantic Taxonomy. \textbf{PS: }Per-part Semantic label. \textbf{PM: }Per-part Mass. \textbf{PI: }Per-part Inertia Moment. \textbf{PF: }Per-part Friction.}
\label{tab:model_comparison}
\end{table*}

\subsection{Fast Articulation Knowledge Modeling (FArM) Pipeline}\label{sec:farm}
Once we determine what to annotate and what object to be annotated, the remaining problem is how to make the annotation process affordable.

\subsubsection{Model Acquisition Equipment.}  To efficiently collect real-world articulated models, we setup a recording system, whose configuration is illustrated in Fig. \ref{fig:model_acq_equipment}. This apparatus is developed with three components: EinScan Pro 2020 for scanning\footnote{\url{https://www.einscan.com}}, Intel RealSense D435 for RGB-D multi-view snapshot, multi-scale rotating turntables and lift bracket. In our setup, each object can be scanned within 5 minutes.

\subsubsection{Articulation Modeling}\label{sec:arti_modeling} After the model acquisition, we develop an articulated object modeling interface with 3D GUI for annotation guidance. Specifically, our modeling workflow split the whole process into three sub-processes:  %The articulation modeling process is illustrated in Fig.\ref{fig:annotation_pipeline}. 

\vspace{0.5em}\noindent\underline{Object Alignment.} This process requires the annotator to align the scanned articulated object from camera space into canonical space which is shared within a category. To assist the alignment, we define several primitive shapes such as cube, sphere and cylinder with predefined axis, which are used to fit the targeted object.

\vspace{0.5em}\noindent\underline{Part Segmentation.} Different from synthetic models from the Internet that often include original mesh subgroups and part information, real-world scanned models require manual segmentation for each rigid part. In our interface, we provide a mesh cutting method with multi-view observation. The annotators draw boundary polygons on the aligned watertight surface and the interface could automatically split the mesh into multiple smaller sub-components. To note, if the parts can be disassembled in the real world, we just scan each part and assemble them into an integral model. 

\vspace{0.5em}\noindent\underline{Joint Annotation.} In contrast to other object modeling pipelines, articulated objects require joint annotation that links two rigid segmented parts and describes the kinematic structure as a tree. Our interface provides an inspector window that allows the annotator to reorganize the parts into a tree structure. Then, the annotators could add joint information to each link and annotate 6D vector (3 for joint location and 3 for joint axis) in a 3D view that contains parent and child parts. To ensure the correctness of joint annotation, we provide an animation that demonstrates the motion under current joint information and the annotators could further refine the annotation.

\begin{figure*}[t]
    \centering
    \includegraphics[width=\linewidth]{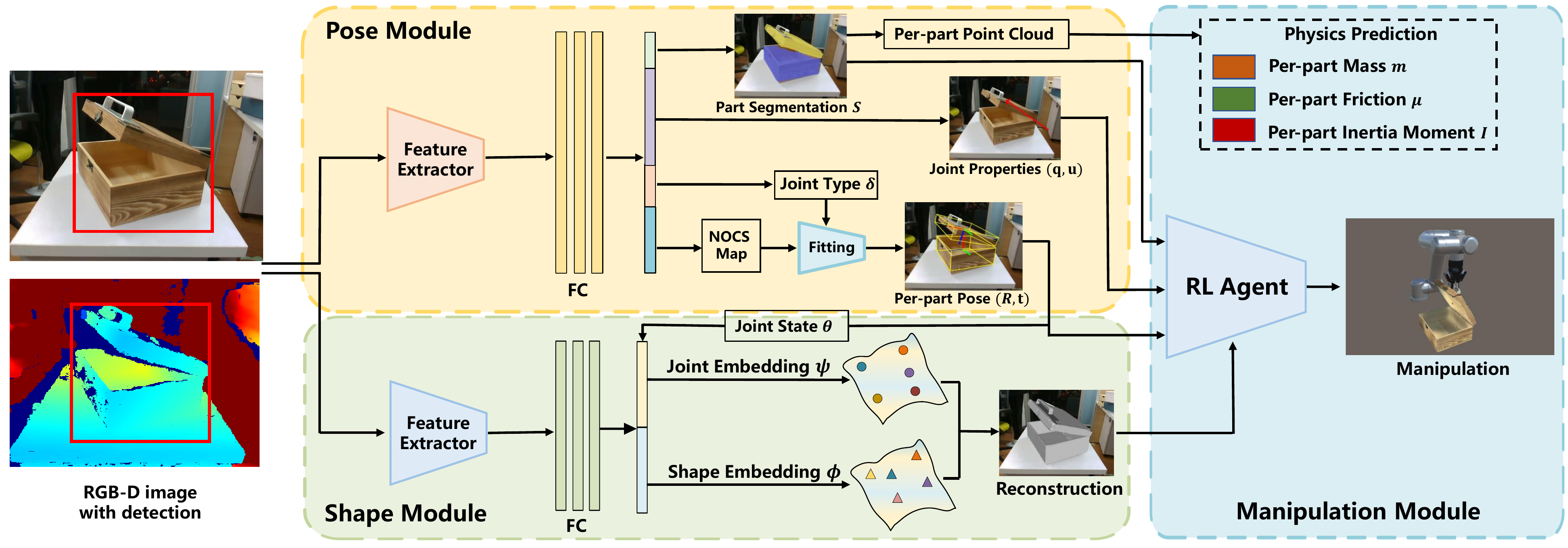}
    %\caption{The baseline method for real-world articulated object pose estimation. There are three modules: object detection, part segmentation \& NOCS map prediction, joint prediction. In this method, we detect each articulated object from RGB-D image and predict each instance's part-level NOCS map with corresponding joint properties.}
    \caption{The overall pipeline of AKBNet. The input of AKBNet is a single RGB-D image with a detected box, and there are three components conducted: (1) Pose module for predicting per-part segmentation, 6D pose, joint type as well as joint properties. (2) Shape module for generating full mesh of the articulated object with current joint state. (3) Manipulation module for enabling the RL agent (UR5 Robot Arm with a Robotiq 85 gripper) to manipulate the object, and also predicting per-part physics information.}
    \label{fig:method}
    \vspace{-0.5cm}
\end{figure*}

\subsubsection{Physics Annotation}\label{sec:phy_ann}

Real-world articulated objects exist in the physical world and have physical properties. To enable our AKB-48 in real-world robotic manipulation \& interaction tasks, we also annotate physical attribute annotations for each part of the articulated object. 

\vspace{0.5em}\noindent\underline{Per-part Mass.} We record each rigid part's weight in grams. For those objects that are inseparable on several parts, we adopt the drainage method \cite{cutnell2014physics} to measure volume for these parts and compute the weight by their densities according to the materials.

\vspace{0.5em}\noindent\underline{Per-part Inertia Moment. } It is hard to obtain per-part inertia moment in the real world since scanned articulated models might contain hundreds of thousands of triangles, which is in a very complicated structure. In our method, we simplify these models with finite primitive shapes, such as cuboid and cone, and then compute the inertial moment in simulation based on the combination of these primitive shapes.

\vspace{0.5em}\noindent\underline{Per-part Material and Friction.} We also annotate the surface material and related parameters. For example, a transparent material will be annotated with the index of refraction, and normal materials will be annotated with friction coefficients. These are obtained by searching Machinery's Handbook \cite{oberg1914machinery}.

\subsection{Dataset Analysis}\label{sec:dataset_analysis}

\noindent\textbf{Object Categories.} To build AKB-48 dataset, we take the following requirements into consideration: (1) \textit{Commonality}. We require our AKB-48 could cover most of the articulated object categories in the common daily scenes, such as kitchen, bedroom and office room. (2) \textit{Variety}. We consider the objects with a wide variety of shapes, deformability, texture and kinematic structure for one category. (3) \textit{Usage}. The chosen objects should contain various functionalities on usage. Besides, the ability to complete manipulation performance is prioritized.

%Current datasets, such as PartNet-Mobility \cite{xiang2020sapien}, usually collect 3D CAD models from Internet. However, most of the models are not suitable for building practical articulation applications {\color{red} such as xx}. We aim to choose objects that are frequently used in daily life, especially in tabletop scenes. Besides, these objects are used for not only real world visual tasks but also manipulation \& interaction tasks in robotics field. Under these requirements, we take the following properties of objects into consideration: (1) \textbf{Commonality}. We require our AKB-48 could cover almost all the common articulated objects in our daily life so we select object categories from several common scenes such as kitchen, bedroom and office room. (2) \textbf{Variety}. To improve the generality of our dataset, we consider the objects with a wide variety of shape, deformability, texture and kinematic structure for one category. (3) \textbf{Usage}. In order to cover as many aspects of robotics manipulation task as possible, the chosen objects should contain various functionalities on how to use them. Besides, the ability to complete manipulation performance is also prioritized.{\color{red} make this paragraph much shorter!}

\vspace{0.5em}\noindent\textbf{Statistics.} We first compare AKB-48 with some other popular datasets in Table \ref{tab:model_comparison}. As it is shown, our object models cover full features for real-world articulated object analysis. Specifically, compared to the synthetic model repository, we hold a much finer surface with average of around 126K triangles and real textures while synthetic models only contain thousands of triangles and synthetic textures. In terms of annotation, we provide part and joint annotations that are enough for visual articulation tasks. Furthermore, we also annotate physical information for each model that is never considered in both synthetic and real-world model repositories before. We believe the rich annotations could promote further development in articulation research. As for the model number, we have a comparable number of objects in comparison with the current largest articulated object datasets PartNet-Mobility\cite{xiang2020sapien}, yet it comprises only CAD models. More statistics such as category specification and intra-category variety can be referred to supplementary materials. 

%To further illustrate the advantages of AKB-48, we investigate the intra-variety in one category of our dataset. We define a shape distribution as metric to measure model shape variety. In detail, we extract Intrinsic Shape Signature (ISS) keypoints \cite{zhong2009intrinsic} from each model. The more ISS keypoints extracted, the more complicated shape is. Given these ISS keypoints, we compute the geometric distance between each keypoint pair, and then do frequency statistics for all the distances \cite{osada2002shape}. Finally, we project these histograms with t-SNE \cite{van2008visualizing}, as shown in Fig. \ref{fig:shape_tsne}. As it could be observed, the models in our AKB-48 hold a large shape variety in one category.

%\begin{figure}[htb]
%    \centering
%    \includegraphics[width=\linewidth]{figures/dataset_analysis3.pdf}
%    \caption{Object shape distribution: Visualization of t-SNE embedding of ISS histograms. A point stands for an instance and a color stands for a category.}
%    \label{fig:shape_tsne}
%\end{figure}

%\subsection{AKB-48-Extend} \label{sec:extend}
%need a demo to compare before/after add rather real property to the objects in simulation.

\section{AKBNet}

In this section, we describe the AKBNet, an integral pipeline for C-VAM task. In AKBNet, the input is a single RGB-D image with detected 2D bounding boxes. We build three sub-modules in AKBNet that aims to estimate per-part 6D pose (Sec. \ref{sec:pose_estimation}), reconstruct full geometry of articulated object (Sec. \ref{sec:object_reconstruction}) and reason the interaction policy through the perception (Sec. \ref{sec:articulation_manipulation}). The overall pipeline of AKBNet is illustrated in Fig. \ref{fig:method}.

%In this section, we describe the AKBNet framework for real-world articulation analysis. In our method, the input is single RGB-D image $\mathcal{I}$ and we firstly employ an object detector to localize each articulated object with the corresponding semantic label. After that, the AKBNet will in parallel predict the per-part information (segments, pose, joint properties, etc.) in a category-level articulation pose estimation task (Sec. \ref{sec:pose_estimation}) and reconstruct the 3D object model in an articulated object reconstruction task (Sec. \ref{sec:object_reconstruction}). Finally, given the visual perceived information, AKBNet will enable an agent to perform articulated object manipulation task (Sec. \ref{sec:articulation_manipulation}). The overall pipeline of AKBNet is illustrated in Fig. \ref{fig:method}.

%In this section, we benchmark three tasks for articulated object analysis. For each of them, we introduce the problem setting and data preparation. We also provide baseline methods for these tasks. 

\subsection{Pose Module}\label{sec:pose_estimation}

Given an image with a detected 2D bounding box, we can obtain the partial point cloud $\mathcal{P}\in \mathbb{R}^{N\times 3}$. Firstly, the input $\mathcal{P}$ is processed by a Pointnet++ \cite{qi2017pointnet++} for feature extraction, and we build two branches at the end for predicting per-point segmentation $S$ and part-level Normalized Object Coordinate Space \cite{li2020category} (NOCS) map $\mathcal{P}'\in\mathbb{R}^{N\times 3}$. To solve the unknown kinematic structure and joint type issues, we introduce three extra branches on the feature extractor to classify the joint type $\delta$ on its corresponding part $k$, and also to predict joint property including joint location $\textbf{q}_i$ and joint axis $\textbf{u}_i$. Finally, we apply the voting scheme to obtain the final joint property $\textbf{q}\in\mathbb{R}^3$ and $\textbf{u}\in\mathbb{R}^3$ . We use cross-entropy loss for part segmentation $\mathcal{L}_{seg}$ and joint type classification $\mathcal{L}_{type}$, L2 loss for NOCS map $\mathcal{L}_{nocs}$, joint location $\mathcal{L}_{loc}$ and joint axis $\mathcal{L}_{ax}$ prediction. Taking all the loss functions into consideration, the overall loss $\mathcal{L}_{pos}$ for pose module is:

\begin{equation}
\begin{split}
\mathcal{L}_{pos} &= \lambda_{seg}\mathcal{L}_{seg} + \lambda_{nocs}\mathcal{L}_{nocs} \\ 
&= + \lambda_{loc}\mathcal{L}_{loc} +  \lambda_{ax}\mathcal{L}_{ax} + \lambda_{type}\mathcal{L}_{type}
\end{split}
\end{equation}

Finally, we follow the pose optimization algorithm with kinematic constrains \cite{li2020category} to recover the 6D pose $\{R,\textbf{t}\}$ for each rigid part. $R$ denotes rotation $R\in SO(3)$ and $\textbf{t}$ denotes translation $\textbf{t}\in\mathbb{R}^3$.

\subsection{Shape Module}\label{sec:object_reconstruction}

Given a partial point cloud $\mathcal{P}$, the shape module aims to re-build the full geometry $\mathcal{M}_{\theta}$ with joint state $\theta$. Followed by A-SDF \cite{mu2021sdf}, we build a feature extractor process the concatenated partial point cloud $\mathcal{P}$ and Gaussian initialized shape embedding $\phi$ as well as joint embedding $\psi$, in which $\phi$ indicates the shape information of the full articulated object and $\psi$ indicates the joint state information that is shared across the same instance. we use SDF values \cite{park2019deepsdf} $d_i$ as supervision and L1 loss for training the shape module $F_{sha}$:

%Given a partial point cloud $\mathcal{P}$, shape module aims to re-build the full mesh $\mathcal{M}$ and generate an sequence of the input object $\{\mathcal{M}_1, \mathcal{M}_2,...,\mathcal{M}_n\}$ with different joint states $\{\theta_1, \theta_2,...,\theta_n\}$. In detail, we also build a feature extractor and SDF \cite{park2019deepsdf} supervision to optimize the shape embedding $\phi$ as well as joint embedding $\psi$, followed by A-SDF \cite{mu2021sdf}: 

\begin{equation}
    \mathcal{L}_{sha} = \lambda_{sha} \frac{1}{N} \sum_{i=1}^N \Vert F_{sha}(p_i,\phi,\psi) - d_i \Vert + \lambda_{\phi} \Vert \phi \Vert_2
\end{equation}

%Given a partial point cloud $\mathcal{P}$ for the detected articulated object, articulated object reconstruction aims to re-build the full mesh $\mathcal{M}$ and generate an sequence of the input object $\{\mathcal{M}_1, \mathcal{M}_2,...,\mathcal{M}_n\}$ with different joint states $\{\theta_1, \theta_2,...,\theta_n\}$. In our AKBNet framework, we provide A-SDF \cite{park2019deepsdf} as baseline method to achieve articulated object reconstruction. In detail, we also build a feature extractor to process the concatenated partial point cloud $\mathcal{P}$ and Gaussian initialized shape embedding $\phi$ as well as joint embedding $\psi$, in which $\phi$ indicates the shape information of the full articulated object and $\psi$ indicates the joint state information that is shared across the same instance. To optimize the shape embedding, joint embedding and model parameters, we use SDF values \cite{park2019deepsdf} $d_i$ as supervision ground truth and L1 loss as the training loss:

During inference, based on the predicted shape embedding $\phi$ and joint embedding $\psi$, we follow Mu's algorithm \cite{park2019deepsdf} to reconstruct the full mesh $\mathcal{M}_{\theta}$. 

%\paragraph{Problem statement.} Given $N$ instance sequences captured from real-world with $M$ joint states, articulated object reconstruction aims to train a model that re-builds the complete point cloud for an input instance sequence and generate the object 3D shapes $\{\mathcal{P}_1, \mathcal{P}_2,...,\mathcal{P}_K\}$ under unseen different joint states $\{\theta_1, \theta_2,...,\theta_K\}$. 

%\paragraph{Data preparation.} We render a large number of single instance images under the same camera configuration with many different joint states. The number of rendered RGB-D images for each instance is 500 (10 joint states plus 50 camera configurations). We also photograph 1000 real images and use the corresponding 3D model for validation.

%\paragraph{Baseline method.} We use A-SDF \cite{mu2021sdf} as the baseline method. In detail, A-SDF trains several Multi Layer Perceptrons (MLPs) to encode the shape and joint state into their feature space. The produced shape and joint state embeddings are then input to an articulated signed distance function to predict SDF values. During inference, A-SDF jointly estimate the shape and joint state given an instance. In addition, new shape with unseen joint states could generated by the trained articulated signed distance function.

\subsection{Manipulation Module}\label{sec:articulation_manipulation}

The manipulation module performs two tasks: opening and pulling that are corresponding to the revolute and prismatic joints in articulation respectively. To achieve these tasks, we train two Reinforcement Learning (RL) agents (UR5 Robot Arm with a Robotiq 85 gripper) these tasks. We provide two \textbf{State Representations:} (1) object state, consisting of 6D pose $\{R, \textbf{t}\}$, joint location $\textbf{q}$, axis $\textbf{u}$, full geometry $\mathcal{M}_\theta$ under current joint state $\theta$. (2) agent state, consisting of the gripper's pose $\{R_{g},\textbf{t}_g\}$ and the gripper's width $w_{g}$. We assume that the agent can access all the information about itself so the agent state is ground truth in our method. The \textbf{Actions} include the agent's end-effector's 3D translation and the opening width of the gripper. The \textbf{Rewards} are rotation angle along the joint axis of the target part for revolute joint and translation distance of that for prismatic joint. The RL agent is trained by two popular RL baselines: Truncated Quantile Critics (TQC)\cite{kuznetsov2020controlling} and Soft Actor-Critic (SAC)\cite{haarnoja2018soft} with Hindsight Experience Replay (HER) \cite{andrychowicz2017hindsight} algorithm.

We also perform physics prediction in our AKBNet. Specifically, the input is a feature vector of point cloud $\mathcal{P}^k$ for $k$th part. We train a 3-layer MLP and build three parallel branches to predict per-part mass $m^k$, friction $\mu^k$ and inertia moment $I^k$. We use L2 loss for training the physics prediction submodule. Please refer to supplementary materials for more details.

\section{Experiments}

\subsection{Experimental Setup}

\noindent\textbf{Dataset.} For the pose module and shape module, we generate 100K RGB-D images with AKB-48 models for training AKBNet using SAMERT data generation scheme \cite{liu2021towards} with scenes from NOCS \cite{wang2019normalized}. And we also capture 10K real-world images, in which 5K are used for fine-tuning the model and the other 5K is test set. For manipulation module, we select 68 and 32 instances for training and testing the RL agent, in which the former is used for opening task and the latter is for pulling task. During training, we use different instances at every episode. 

\vspace{0.5em}\noindent\textbf{Implementation Details.} When training pose module and shape module, we use Adam optimizer with initial learning rate 0.001. Batch size is 16. The total training epochs are 50 and 100 for training these two modules. The detailed hyper-parameters are: $\lambda_{seg}=1$, $\lambda_{nocs}=10$, $\lambda_{loc}=1$, $\lambda_{ax}=0.5$, $\lambda_{type}=1$, $\lambda_{sha}=1$, $\lambda_{\phi}=0.0001$. For the manipulation module, the hyper-parameters are: batch size is 512, learning rate is 0.001, replay buffer size is 100K, soft update coefficient is 0.05, discount factor is 0.95. We use RFUniverse \cite{rfuniverse} as the environment to train the RL agent. For more details, please refer to the supplementary materials.

%we evaluate the performance on 60 instances that contain revolute joints for supporting opening task and 30 instances that contain prismatic joints for support pulling task. The total training step is 100K. When evaluating manipulation module, we regard visual states (image and point cloud) as raw input, and investigate the effects of physics as well as object state prediction. For more details, please refer to the supplementary materials.

\vspace{0.5em}\noindent\textbf{Metrics.} We adopt the following metrics to measure the AKBNet performance. For the pose module, We report three part-based metrics: rotation error measured in degrees, translation error measured in meters and 3D IoU for each part. We also report the joint-based metrics: angle error of joint axis measured in degrees, location error in line-to-line distance measured in meters, joint type classification accuracy (\%). For the shape module, we report the average Chamfer-L1 distance \cite{mu2021sdf} for reconstruction evaluation. For the manipulation module, we report success rate (\%) as the metric. If the agent can grip the target part and move it through 50\% of its motion range, it will be regarded as a success.

%We apply sim-to-real scheme to train AKBNet on synthetic dataset and validate the performance on real-world data, in which we synthesize 100K images for training and capture 10K for testing. The input images are resized into (1333, 800) for better detection performance referred by \cite{lin2014microsoft} and we choose RetinaNet \cite{lin2017focal} with ResNet-50 \cite{he2016deep} backbone as detector. For category-level articulation pose estimation and articulated object reconstruction task, we use Adam optimizer with initial learning rate 0.001. The total training epochs are 50 and 100 for the two tasks. The detailed hyper-parameters are: $\lambda_{seg}=1$, $\lambda_{nocs}=10$, $\lambda_{loc}=1$, $\lambda_{ax}=0.5$, $\lambda_{type}=1$, $\lambda_{rec}=1$. For articulated object manipulation, the total training step is 100K. 

\begin{figure*}[t]
    \centering
    \includegraphics[width=\linewidth]{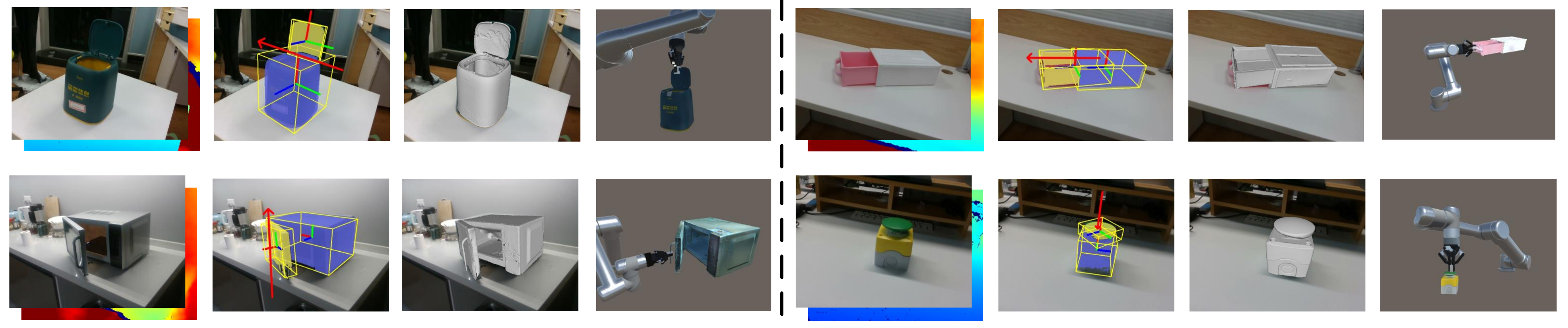}
    \caption{Qualitative results. For one instance, from left to right: input RGB-D image, output of pose module, output of shape module, manipulation demonstration.}
    \label{fig:qualitative_res}
    \vspace{-0.5cm}
\end{figure*}

\subsection{Pose Module Performance}

We evaluate NPCS \cite{li2020category}, A-NCSH \cite{li2020category} and AKBNet on real-world test set for category-level articulation pose estimation task. For A-NCSH baseline, we use direct regression and classification scheme to predict kinematic structure and joint type. The experimental results are illustrated in Table \ref{tab:res_pose_estimation}. For pose estimation, we achieve \textbf{9.8}, \textbf{0.021} and \textbf{53.6} on rotation, translation errors and 3D IoU, which are higher than NPCS and A-NCSH. For joint-related evaluation, we can precisely predict joint type for unseen articulated objects with \textbf{94.6\%} accuracy. Besides, AKBNet achieves \textbf{8.1} and \textbf{0.019} errors in joint axis and location prediction respectively.

%In comparison with A-NCSH, we can also obtain comparable performance without ground truth kinematic structure and joint type for inference. For joint-related evaluation, we can precisely predict joint type for unseen articulated objects with \textbf{94.6\%} accuracy. Besides, AKBNet achieves \textbf{8.1} and \textbf{0.019} errors in joint properties prediction. 

\begin{table}[thb]
\scriptsize
\small
\centering
\resizebox{0.9\linewidth}{!}{
\begin{tabular}{c|ccc}
\hline
\multirow{2}{*}{Method} & \multicolumn{3}{c}{Part-based Metrics} \\
\cline{2-4}
& rotation$\downarrow$ & translation$\downarrow$ & 3D IoU$\uparrow$ \\
%Method & Rotation error $\downarrow$ & Translation error $\downarrow$ & 3D IoU $\uparrow$ \\
% \multirow{2}{*}{Method} & \multicolumn{2}{c}{Rotation} & \multicolumn{2}{|c}{Translation} & 
% \multicolumn{2}{|c}{3D IoU} \\
% \cline{2-7}
%  &  5$^\circ$ & 10$^\circ$ & 5cm & 10cm & 0.5 & 0.7 \\
\hline
% NPCS\cite{li2020category} & 20.5 & 46.9 & 51.2 & 80.3 & 22.8 & 8.6 \\
% A-NCSH*\cite{li2020category} & 23.1 & 49.3 & 55.8 & 85.6 & 26.5 & 10.9 \\
% AKBNet & \textbf{21.8} & \textbf{49.0} & \textbf{54.6} & \textbf{85.3} & \textbf{25.8} & \textbf{10.5} \\
NPCS\cite{li2020category} & 12.6 & 0.038 & 48.3 \\
A-NCSH*\cite{li2020category} & 10.5 & 0.026 & 50.8 \\
AKBNet & \textbf{9.8} & \textbf{0.021} & \textbf{53.6} \\
\hline
\multirow{2}{*}{Method} & \multicolumn{3}{c}{Joint-based Metrics} \\
\cline{2-4}
& angle$\downarrow$ & distance$\downarrow$ & type$\uparrow$ \\
\hline
NPCS\cite{li2020category} & - & - & - \\
A-NCSH*\cite{li2020category} & 9.2 & 0.021 & 93.8 \\
AKBNet & \textbf{8.1} & \textbf{0.019} & \textbf{94.6} \\
\hline
\end{tabular}}
\caption{Category-level articulation pose estimation results. $\downarrow$ means the lower the better. $\uparrow$ means the higher the better. * indicates that A-NCSH is re-implemented with the extra kinematic structure and joint type prediction modules.}
\label{tab:res_pose_estimation}
\vspace{-0.2cm}
\end{table}

\begin{comment}
\subsection{Joint Estimation Performance} 

Since category-independent articulation joint estimation is a new task and no current methods target at it, we evaluate our proposed method and use average angle error and distance error as metrics. We also compare the performance with the ablated versions of our method. The experimental results are shown in Table \ref{tab:res_joint_estimation}.

\begin{table}[thb]
\scriptsize
\small
\centering
\resizebox{\linewidth}{!}{
\begin{tabular}{c|cc|c}
\hline
\multirow{2}{*}{Method} & \multicolumn{2}{c}{Revolute Joint} & \multicolumn{1}{|c}{Prismatic Joint} \\
\cline{2-4}
 &  Angle Error & Distance error & Angle error \\
\hline
Ours+G & 11.2$^\circ$ & 3.4cm & 9.6$^\circ$ \\
Ours+K  & 18.6$^\circ$ & 9.7cm & 18.3$^\circ$ \\
Ours+G+K  & 9.4$^\circ$ & 2.8cm & 7.5$^\circ$ \\
\hline
\end{tabular}}
\caption{Category-independent articulation joint estimation results. We report joint angle error and distance error for revoulte joint, and only angle error for prismatic joint. G indicates Geometric constrains and K indicates Kinematic constrains during joint parameter optimization.}
\label{tab:res_joint_estimation}
\end{table}
\end{comment}

\subsection{Shape Module Performance} 

The experimental results of the shape module are illustrated in Table \ref{tab:res_reconstruction}. Within ground truth joint state input, the shape module could reconstruct the articulated object with \textbf{4.2} Chamfer-L1 distance. On the other hand, we systematically evaluate the shape module given the predicted joint state, which is deduced from predicted the linked two parts' poses from the pose module. The Chamfer-L1 distance is \textbf{3.3} higher than that with ground truth joint state, indicating that the predicated poses largely affect reconstruction performance.

\begin{table}[thb]
\scriptsize
\small
\centering
\resizebox{0.7\linewidth}{!}{
\begin{tabular}{l|cc}
\hline
Mode & Chamer-L1 Distance \\
\hline
Joint State GT & 4.2 \\
Joint State Pre. & 7.5 \\
\hline
\end{tabular}}
\caption{Articulated object reconstruction results. Pre. means that we use the predicted joint state from the pose module.}
\label{tab:res_reconstruction}
\vspace{-0.5cm}
\end{table}

\subsection{Manipulation Module Performance}

We evaluate opening and pulling tasks on the manipulation module of AKBNet using TQC+HER training algorithm compared with that using SAC+HER. Experimental results are illustrated in Table \ref{tab:res_manipulation}. With ground truth object state, AKBNet could complete opening and pulling manipulation tasks, with \textbf{72.5\%} and \textbf{98.7\%} success rate. However, our method might not perform well when the object state is predicted, with only 40.2\% and 44.6\% success rates. Qualitative results of AKBNet are illustrated in Fig. \ref{fig:qualitative_res}.

Our AKBNet can also predict physics information including per-part mass, friction and inertia moment. These predicted physics can enable force sensing for AKB-48 objects in simulation, which has the potential to realize force controlling. For more details, please refer to supplementary materials. 
%which indicates that our defined ArtiKG is essential for articulated object interaction \& manipulation. 

%We evaluate SAC \cite{haarnoja2018soft} to train a reinforcement learning agent on box opening and drawer pulling task. We test the agent on synthetic as well as real-world environment and use manipulation success rate as evaluation metric. The experimental results are shown in Table \ref{tab:res_manipulation}.

\begin{table}[thb]
\scriptsize
\small
\centering
\resizebox{\linewidth}{!}{
\begin{tabular}{c|c|c|c}
\hline
Method & Mode & Opening & Pulling \\
%\multirow{2}{*}{Mode} & \multicolumn{2}{c}{Task} \\
%\cmidrule(r){2-3}
%& Opening & Pulling \\
\hline
\multirow{2}{*}{AKBNet+SAC\cite{haarnoja2018soft}+HER\cite{andrychowicz2017hindsight}} & Object State GT & 57.1 & \textbf{98.7} \\
& Object State Pre. & 32.3 & 36.5 \\
\hline
\multirow{2}{*}{AKBNet+TQC\cite{kuznetsov2020controlling}+HER\cite{andrychowicz2017hindsight}} & Object State GT & \textbf{72.5} & 95.5 \\
& Object State Pre. & \textbf{40.2} & \textbf{44.6} \\
\hline
\end{tabular}}
\caption{Success rate (\%) on articulated object manipulation task. Pre. means we use predicted object state from the pose and shape modules.}
\label{tab:res_manipulation}
\vspace{-0.5cm}
\end{table}

\section{Conclusion and Crowd-Sourcing Data-Collection Invitation}

In this paper, we present AKB-48, a large-scale articulated object knowledge and benchmark C-VAM task for dealing with articulation problems. Admittedly, there are a few articulated object categories that might not be collected in AKB-48, although we have covered large enough categories in daily life. In the future, we will release our FArM tool for collecting more articulated objects, and it could also support any scanned shapes such as mobile reconstructor \cite{klingensmith2015chisel}. In future work, we will publish an online articulation model platform and invite crowd-sourcing data-collection to contribute to the articulation research community.

%AKB-48 covers large enough articulation categories in our daily life, but there are a few categories that might not be collected. In the future, we will release our FArM tool for collecting more articulated objects, and it could also support any scanned shapes such as mobile reconstructor \cite{klingensmith2015chisel}. In future work, we will publish an online articulation model platform and invite crowd-sourcing data-collection to contribute to the articulation research community.

%Indeed, we notice that AKB-48 might not contain a large amount of expensive objects. However, we will release our FArM tool for collecting arbitrary articulated objects, and it could also support any scanned shapes such as mobile reconstructor \cite{klingensmith2015chisel}. In the future work, we will publish an online articulation model platform and invite crowd-sourcing data-collection to contribute to the articulation research community.

% We present AKB-48: a large-scale articulated object knowledge base for bridging the gap between simulation and real-world articulation analysis. It contains 2,037 real-world 3D articulated object models of 48 categories. To efficiently build AKB-48, we propose a fast articulation object modeling pipeline that achieves rich articulation knowledge annotation within 10-15 minutes. We benchmark four tasks for real-world articulation analysis evaluation, including category-level pose estimation, category-independent joint estimation, articulation reconstruction and articulation object manipulation.

%%%%%%%%% REFERENCES
{\small
\bibliographystyle{ieee_fullname}
\bibliography{egbib}
}

\end{document}